# House Price Prediction Using LSTM

Xiaochen Chen          Lai Wei          Jiaxin Xu

The Hong Kong University of
Science and Technology

## ABSTRACT
In this paper, we use the house price data ranging from January 2004 to October 2016 to predict the average house price of November and December in 2016 for each district in Beijing, Shanghai, Guangzhou and Shenzhen. We apply Autoregressive Integrated Moving Average model to generate the baseline while LSTM networks to build prediction model. These algorithms are compared in terms of Mean Squared Error. The result shows that the LSTM model has excellent properties with respect to predict time series. Also, stateful LSTM networks and stack LSTM networks are employed to further study the improvement of accuracy of the house prediction model.

## 1. INTRODUCTION
House price plays a significant role in shaping the economy. Housing renovation and construction boost the economy by increasing the house sales rate, employment and expenditures. The traditional tedious price prediction process is based on the sales price comparison and stochastic process prediction. which hardly achieve valuable accuracy. Therefore, we aim to find a more reliable way to predict the house price trend based on RNN model.

This study focuses on the housing market in Beijing, Shanghai, Guangzhou, Shenzhen. We collected the house price information. After observation, we find location and time are the main factors of determine real estate prices. Therefore, we aggregate our data according to districts.

## 2. Design and Implementation
### 2.1 Overview Solution
We use a machine learning model named RNN (Recurrent Neural Network) as the foundation of our solution. LSTM(Long Short Term Memory) is a variation of RNN, which has successfully tackled a lot problem in AI. We intend to use LSTM as backbone of our solution.

### 2.2 Preprocessing
We collected the house price data together with the house attributes ranging from 2004 to 2016. However, there is some missing and not applicable attribute values. Besides, some attributes which are redundant or not suitable for computing need to be unified. Considering the cost will be high resulting from the large data amount, we decide to use the average house price of each district. The model we use only counting the house price attribute and time attribute. As above, we use three steps to preprocess our data. First, eliminate tuples with missing house price. Second, unify the format of time attribute. Third, computing the average house price of each district.

### 2.3 Baseline System
We use ARIMA as our baseline System. ARIMA is a mature time series prediction model based on statistics. Which is widely used in the field of High Frequency Trade. It can be adapted to the house price prediction task as it's a general time series model.

Our perspective is to build up better prediction using machine learning. ARIMA as a non-learning based method, performs well as a baseline.

### 2.4 RNN
A recurrent neural network (RNN) is a class of artificial neural network where connections between units form a directed cycle. RNNs can use their internal memory to process arbitrary sequences of inputs.

There are a variety types of RNN and we only consider the Simple Recurrent Network(SRN) and LSTM network. We will use Mean Squared Error(MSE) to compare the accuracy of the prediction model constructed by these two networks.

The basic architecture was employed by Jeff Elman[4]. A three-layer network is used with the addition of a set of "context units". There are connections from the middle (hidden) layer to these context units fixed with a weight of one.[5] At each time step, the input is propagated in a standard feed-forward fashion, and then a learning rule is applied. The fixed back connections result in the context units always maintaining a copy of the previous values of the hidden units (since they propagate over the connections before the learning rule is applied). Hence, the network can maintain a sort of states, allowing it to perform such tasks as sequence-prediction that are beyond the power of a standard multilayer perceptron.

$$h_t = \sigma_h(W_h x_t + U_h h_{t-1} + b_h)$$
$$y_t = \sigma_y(W_y h_t + b_y)$$

Variables and functions:

$x_t$ : $input\ vector$

$h_t$: $hidden\ layer\ vector$

$y_t$: $output\ vector$

W, U and b: parameter matrices and vector

$\sigma_h$ and $\sigma_y$: $Activation\ functions$

### 2.5 LSTM
The simple RNN system is not good enough to do the prediction, so we decide to use the more complicated LSTM architecture in RNN system. LSTM is shorted for "Long short-term memory".

A LSTM network is an artificial neural network that contains LSTM units instead of, or in addition to, other network units. A LSTM unit is a recurrent network unit that excellent at remembering values for either long or short durations of time. The key to this ability is that it uses no activation function within its recurrent components. Thus, the stored value is not iteratively squashed over time, and the gradient or blame term does not tend to vanish when Backpropagation through time is applied to train it.

LSTM units are often implemented in "blocks" containing several LSTM units. This design is typical with "deep" multi-layered neural networks, and facilitates implementations with parallel hardware. In the equations below, each variable in lowercase italics represents a vector with a size equal to the number of LSTM units in the block.

LSTM blocks contain three or four "gates" to control the flow of information into or out of their memory. These gates are implemented using the logistic function to compute a value between 0 and 1. Multiplication is applied with this value to partially allow or deny information to flow into or out of the memory. For example, an "input gate" controls the extent to which a new value flows into the memory. A "forget gate" controls the extent to which a value remains in memory. And, an "output gate" controls the extent to which the value in memory is used to compute the output activation of the block. (In some implementations, the input gate and forget gate are combined into a single gate. The intuition for combining them is that the time to forget is when a new value worth remembering becomes available.)

The only weights in a LSTM block (W and U) are used to direct the operation of the gates. These weights occur between the values that feed into the block (including the input vector $x_t$, and the output from the previous time step $h_{t-1}$) and each of the gates. Thus, the LSTM block determines how to maintain its memory as a function of those values, and training its weights causes the LSTM block to learn the function that minimizes loss. LSTM blocks are usually trained with Backpropagation through time.

*2.5.1 Simple Model*

This picture below shows the simple model of LSTM. Every time the model uses five in order records of data to be the input and finally produce one output. This output value will affect the parameters of the LSTM system for the next iteration. So that it can improve the simulation with every adjustion.

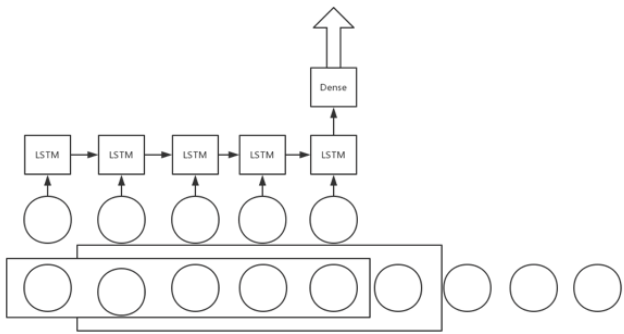

Figure 1: simple model of LSTM

Initial values: $c_0 = 0$ and $h_0 = 0$. The operator ∘ denotes the Hadamard product (entry-wise product).[6][7]

$$f_t = \sigma_g(W_f x_t + U_f h_{t-1} + b_f)$$
$$i_t = \sigma_g(W_i x_t + U_i h_{t-1} + b_i)$$
$$o_t = \sigma_g(W_o x_t + U_o h_{t-1} + b_o)$$
$$c_t = f_t \circ c_{t-1} + i_t \circ \sigma_c(W_c x_t + U_c h_{t-1} + c)$$
$$h_t = o_t \circ \sigma_h(c_t)$$

Variables:

$x_t$: input vector

$h_t$: output vector

$c_t$: cell state vector

W,U and b: parameter matrices and vector

$f_t, i_t, o_t$: gate vectors

$f_t$: Forget gate vector. Weight of remembering old information.

$i_t$: Input gate vector. Weight of acquiring new information.

$o_t$: Output gate vector. Output candidate.

Activation functions:

$\sigma_g$: The original is a sigmoid function.

$\sigma_c$: The original is a hyperbolic tangent.

$\sigma_h$: The original is a hyperbolic tangent

## 2.6 Stateful LSTM

In Recurrent Neural Networks, we are quickly confronted to the so-called gradient vanishing problem:

In machine learning, the vanishing gradient problem is a difficulty found in training artificial neural networks with gradient-based learning methods and backpropagation. In such methods, each of the neural network's weights receives an update proportional to the gradient of the error function with respect to the current weight in each iteration of training. Traditional activation functions such as the hyperbolic tangent function have gradients in the range $(-1,1)$ or $[0,1)$, and backpropagation computes gradients by the chain rule. This has the effect of multiplying n of these small numbers to compute gradients of the "front" layers in an n-layer network, meaning that the gradient (error signal) decreases exponentially with n and the front layers train very slowly.

One solution is to consider adding the updates instead of multiplying them, and this is exactly what the LSTM does.

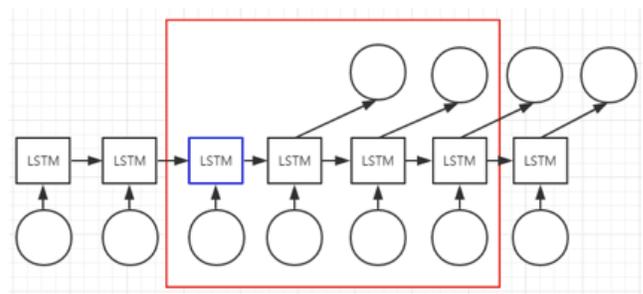

Figure 2: simple model of Stateful LSTM

Figure 2 shows Stateful LSTM architecture, When we train time steps in red rectangle, the start time step is blue one. The blue one Initial values: $c_0$ and $h_0$ is not 0. It's value is trained from the 0 time steps.

## 2.7 Stacked LSTM

Stacked LSTM is a variation of LSTM inspired by Deep Learning.

Which uses more than one LSTM layer to build up the Neural Network. Thus, able to model much more complex underlying data trends. But also require bigger dataset to train.

### 2.7.1 The architecture

In our experiment. We used 3 layers LSTM, each layer has size of 64 hidden units. The first two layer were sequence to sequence layer and the third layer is sequence to single output. Followed by a fully connected dense layer

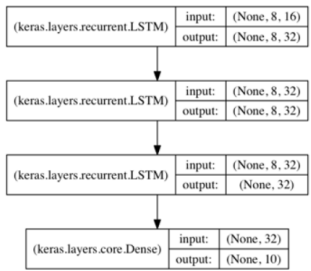

Figure 3: simple model of Stacked LSTM

## 3. Evaluation

### 3.1 Dataset

Our dataset has house price in Beijing, Shanghai, Guangzhou and Shenzhen from 2004.1 to 2016.9. Including 80 districts from or near these 4 cities. We are going to predict the average house price (RMB/$m^2$) of 2016.10 and 2016.11 for each district in Beijing, Shanghai, Guangzhou, and Shenzhen.

We use python to group data into month and district. Then calculate average price. Then we can have average price in different district in 154 months.

In our experiments except ARIMA model, we use 15 months to predict 1 month, so we can have 139 data. We split 139 data into 14 validation data and 125 training data using cross validation convention.

### 3.2 ARIMA model

We use a python open source package called Pyflux to implement our ARIMA model. Pyflux package is developed by Pydata. We firstly change our price to return, which is commonly preprocess in financial area. Then we chose ar=4, ma=4 as our ARIMA model parameter. Using Maximum Likelihood Estimation to fit our return rate to ARIMA model. In our experiment, we use the first 140 month to train, and the last 14 month to test our model accuracy. Here we only consider Chaoyang district from Beijing.

Figure 1 is our model training result.

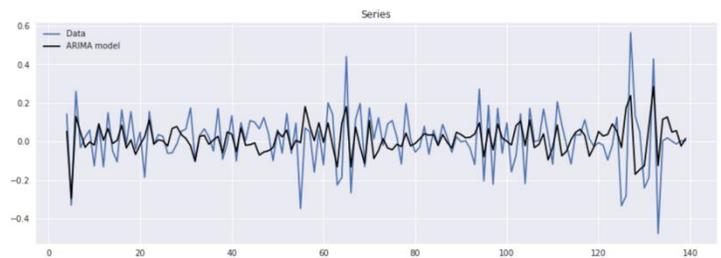

Figure 4: ARIMA training result

Figure 2 shows our ARIMA model and compare to the ground truth of return rate.

Figure 5: ARIMA model and ground truth

Then we try to predict the next 14 months. And then change predicted return rate to original price. Because we need to compare for else method, so we normalize price to [0,1] and compute Mean Squared Error on these 14 months. Our ARIMA model MSE on training set is 0.0389.

### 3.3 Simple RNN model

In our experiment, we use a single hidden layer RNN with only 4 hidden units. Use 15 days to predict 1 day. In each training epoch, we go through all the 155 month. Each time. Figure 6 shows the basic structure of our RNN model. Here we only consider Chaoyang district from Beijing.

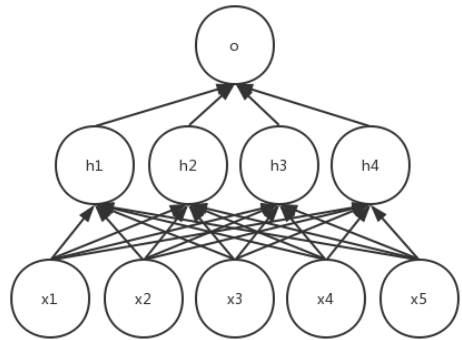

Figure 6: Simple RNN model

This RNN has 5 units to input. X1 is current time average. In the first month, x2 to x5 is 0, in the following time steps, x2 to x5 is equal to h1 to h4 values in the former time step. After 15 iteration, we get an output, and use this output as our predict.

Figure 7 shows comparison between our model fit data and ground truth.

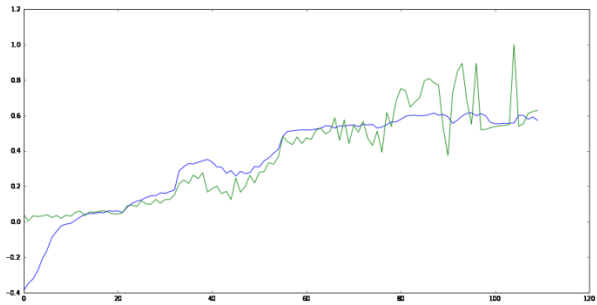

Figure 7: simple RNN and ground truth

Figure 8 shows our training converge progress.

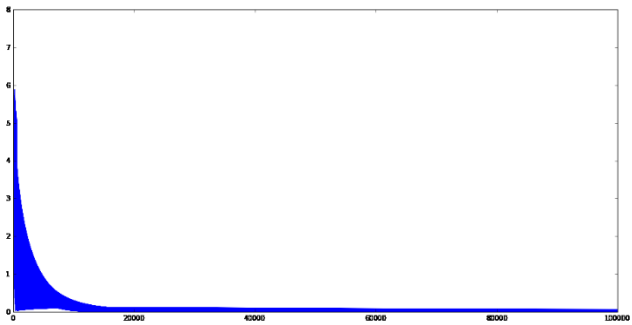

Figure 8: Simple RNN converge

After training, we get a MSE on validation set, which is 0.018970.

### 3.4 LSTM model

Here we only consider Chaoyang district from Beijing. We use a single layer LSTM with 256 hidden units. Input is average price of current time step, only one dimension. Output is also average price of next time step, only one dimension. Each time we input current average price, after 15 times, we calculate the output of the last batch, use this output as our predict of next average price of next month.

Figure 9 shows the comparison between model prediction and ground truth.

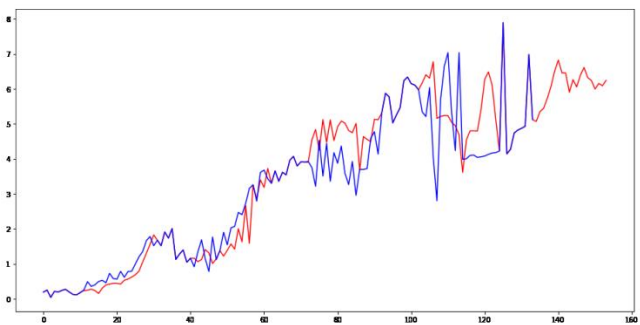

Figure 9: LSTM and ground truth

During our training, every 100 batches we record MSE of current batch and calculate the MSE on the validation set. Figure 7 shows this two values converge during our training.

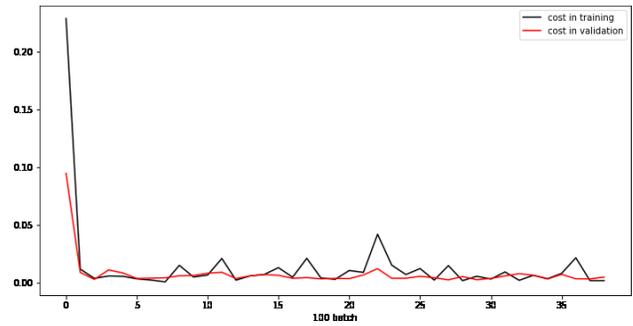

Figure 10: training MSE and validation MSE

### 3.5 Combine all District LSTM model

We plug our LSTM in 3.1.4 into all district. So the input is not only 1 dimension. It is 80 dimension. The input of our model is 80 average price over all districts in current month, and Output is 80 average price over all district in the next month.

We first one hot encoding our dataset, make it into 155*80 matrix named $dataMtx$. Each row represents one month, each column represents one district. Then make training data one by one. For example,

$$\chi^0 = dataMtx[0:15,:]$$
$$Y^0 = dataMtx[15,:]$$

We regard each training data is independent. Each time we chose training batch randomly.

During training, every 100 batches we record MSE on current batch and MSE on validation set. Figure 8 shows this two values.

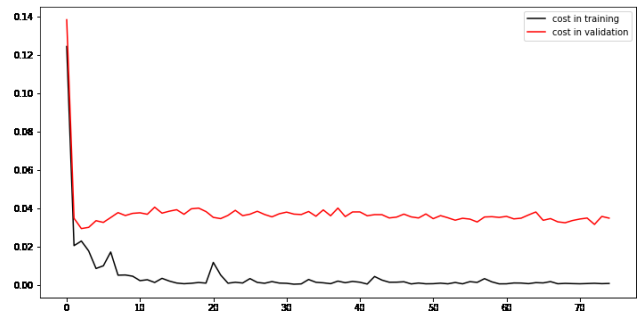

Figure 11: training MSE and validation MSE

We can see after combine all district, the MSE in validation increased dramatically.

In order to know which is the best number of hidden units. We did series of experiment. Figure 9 shows MSE of different units.

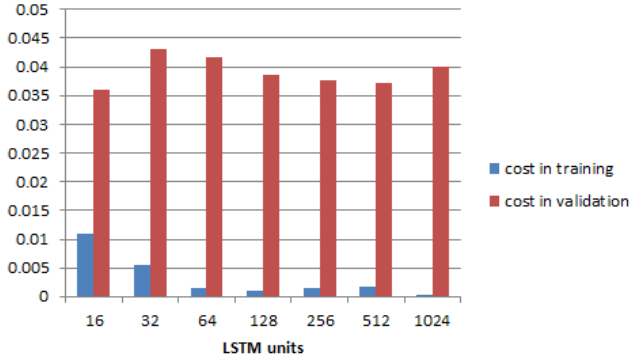

Figure 12: different number of LSTM units

From figure 12 we can see, when number of units increase, model did better in training set, fits training set better, but almost no difference in validation set.

### 3.6 Stateful LSTM model

In stateful LSTM model, we only use the first 135 training data, reshape it into 5*27 matrix called batchMtx, 5 is our batch size,

$$\text{Training data} = \text{batchMtx}[:, 0:25, :]$$

$$\text{Test data} = \text{batchMtx}[:, 25:27, :]$$

We do not chose training batch randomly each time. We train our training data sequentially. In non-stateful model, the first month is empty LSTM, but in stateful model, the first month is the second month in the former batch. That means the first month use the value of the second month in the former batch to initialize. So now our first month also have the memory. Figure 10 shows when we use this model, MSE over each training batch.

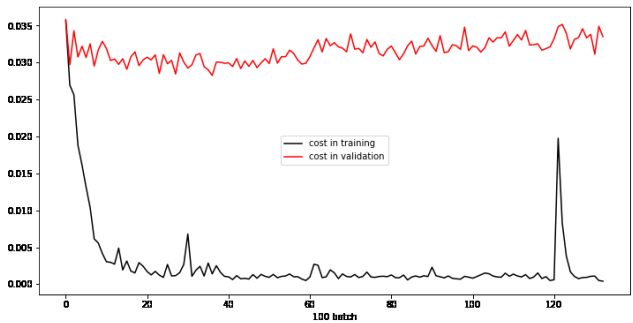

Figure 13: training MSE and validation MSE

The experiment demonstrated that this model doesn't work well on the validation set. It's a little overfitted.

### 3.7 Stacked LSTM

In stacked LSTM model, we used 3-layer LSTM, each layer have size 64 hidden units.

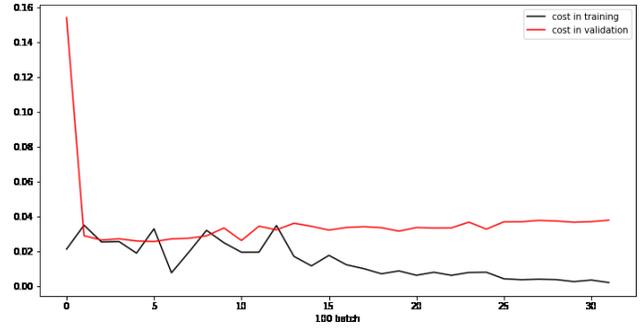

Figure 14: training MSE and validation MSE

## 4. Conclusion

We use the house price information from 80 districts in 155 months to build up the prediction model and made good prediction in next 2 months. The noticeable improvement has been achieved.

### 4.1 baseline

The prediction is much better than baseline ARIMA, about 90 percent reduce on MSE.

### 4.2 stateful LSTM

The stateful LSTM model has bad prediction on validation data set. Which indicate that the essence of the problem may not suitable for the application of stateful LSTM.

### 4.3 stacked LSTM

The stacked LSTM has similar accuracy with Basic LSTM. In fact, deeper neural network is expected to give a better result than the simple one, which was not seen in evaluation. Which inspire us to do further exploration on finding better structure and parameter for stacked LSTM and improve the accuracy.

### 4.4 The problems

The result in some districts are not ideal. The possible reasons are low frequency of data and the loss of data in several months.

## 5. Future Work

As we found several problems. In future, we will do more research on stacked LSTM, which is just simply implemented using fixed structure and parameters. Given more time, the even comprehensive study on stacked LSTM may be done. At the same time. Poor data quality and amount limited the accuracy of the model. By introducing bigger dataset and handle the data smarter, the prediction may be improved dramatically.